\documentclass{article}
\usepackage{ijcai23}

\usepackage{times}
\usepackage{soul}
\usepackage{url}
\usepackage[hidelinks]{hyperref}
\usepackage[utf8]{inputenc}
\usepackage[small]{caption}
\usepackage{graphicx}
\usepackage{amsmath}
\usepackage{amsthm}
\usepackage{booktabs}
\usepackage{algorithm}
\usepackage[switch]{lineno}

\usepackage{multirow}
\usepackage{booktabs}

\usepackage{amssymb}
\usepackage{pifont}

\title{A Survey on Out-of-Distribution Evaluation of Neural NLP Models}

\author{
Xinzhe Li$^1$
\and
Ming Liu$^1$
\and
Shang Gao$^{1}$\And
Wray Buntine$^2$
\affiliations
$^1$Deakin University\\
$^2$VinUniversity
\emails
\{lixinzhe, m.liu,shang.gao\}@deakin.edu.au, wray.b@vinuni.edu.vn
}

\begin{document}

\maketitle

\begin{abstract}
    Adversarial robustness, domain generalization and dataset biases are three active lines of research contributing to out-of-distribution (OOD) evaluation on neural NLP models.
    However, a comprehensive, integrated discussion of the three research lines is still lacking in the literature. In this survey, we 
    1) compare the three lines of research under a unifying definition; 
    2) summarize the data-generating processes and evaluation protocols for each line of research; and 3) emphasize the challenges and opportunities for future work.
\end{abstract}

\section{Introduction}
Neural NLP models have been reported for their superhuman performance on many language understanding and generation tasks, such as sentiment analysis and machine reading comprehension (MRC). 
Recent studies show that these models lack human-level understanding of language since they are vulnerable beyond in-distribution (ID) test data and fail to generalize on out-of-distribution (OOD) data, such as perturbed examples under adversarial attacks \citep{ebrahimi2017hotflip}, text from different domains \citep{hendrycks-etal-2020-pretrained} and examples against dataset biases \citep{gururangan2018annotation}, which lie in the three mainstream lines of research: adversarial robustness, domain generalization and dataset biases.

Despite recent progress in each of the three research lines, there has not been a survey that comprehensively reviews and investigates the connections between 
 these lines. 
We have the following motivations for such an investigation:  1) explaining some phenomena across the three lines of research. For example, methods designed to improve model generalization on one OOD type can defend against other types of OOD data \citep{pmlr-v139-yi21a} but sometimes degrade model performance on other OOD types \citep{pmlr-v119-bras20a,gokhale-etal-2022-generalized}; 
2) encouraging future researchers to perform a comprehensive OOD evaluation while proposing their new methods; 
3) suggesting that all the OOD types be addressed thoroughly in future research.

In this survey, we first unify the three research lines as the study of distribution shift, 
which gives us a starting point to find connections between them in terms of shifted features (\S \ref{sec:define}).
Furthermore, we summarize their differences in OOD data generation and evaluation protocols (\S \ref{sec:evaluate_ood}). In particular, we categorize different methods to generate OOD data, including data with natural domain shift (NDS) for domain generalization, debiased data for evaluating dataset biases and adversarial examples for adversarial robustness. We then categorize evaluation protocols for the OOD data into two classes: data-based and method-based evaluation.
Finally, we identify opportunities based on the discussion of connections and differences between the three OOD types, including developing a comprehensive benchmark for OOD evaluation, advising caution about breaking the assumptions of covariate shift, improving general OOD performance and using the detection methods across different OOD types. Also, we demonstrate the gap in adversarial robustness for future work (\S \ref{sec:challenges}).

\section{Definition}  \label{sec:define}
This section first introduces a comprehensive definition of distribution shift across the three research lines and then shows how the definition aids in elucidating the interconnections regarding the shifted features. 

\subsection{Distribution Shift}
Deep Neural Network (DNN) $f(\theta)$ can achieve remarkable or even superhuman performance on NLP benchmarks, such as GLUE with in-distribution (ID) test data $\mathcal{D}{\text{test}} \sim \mathbb{P}_0$, when $f(\theta)$ is trained on data sampled from the data-generating distribution (an unknown distribution under a data-generating process) $\mathcal{D}_{\text{train}} \sim \mathbb{P}_0$. However, the well-trained model often fails to generalize to various unforeseen examples.
Formally, the unforeseen data (or OOD data) $\Tilde{ \mathcal{D}}$ can be characterized by a shifted distribution $\Tilde{\mathbb{P}}$. 
The distribution shift includes covariate shifts (input distribution shifts) $\mathbb{P}_0(\mathcal{X}) \neq \Tilde{\mathbb{P}}(\mathcal{X})$ and label distribution shifts $\mathbb{P}_0(\mathcal{Y}) \neq \Tilde{\mathbb{P}}(\mathcal{Y})$, where $\mathcal{X}$ is the input space, and $\mathcal{Y}$ is the ground-truth output space. We unify the three lines of research as a study of covariate shifts. For brevity, the input notation $\mathcal{X}$ is sometimes omitted when referring to it, as exemplified by the use of $\mathbb{P}_0$ to represent $\mathbb{P}_0(\mathcal{X})$.

\paragraph{Domain generalization and dataset biases.}
Real-life documents in different domains, e.g., news and fiction text \citep{hendrycks-etal-2020-pretrained}, are typically characterized by systematically distinct properties and originate in different data-generating distributions.
Domain generalization investigates such natural domain shifts, and relevant literature often uses domain shift and distribution shift interchangeably.
Some literature \citep{arora-etal-2021-types,gokhale-etal-2022-generalized} also defines that data with particular biases and those without such biases come from different domains. Therefore, the studies of domain generalization and dataset biases can be intuitively defined in the framework of distribution shifts.

\paragraph{Adversarial robustness: from robustness to distribution shift.}
Initially, adversarial methods evaluate model robustness against small perturbations in the worst-case scenarios, especially for continuous data like images \citep{goodfellow2014explaining}.
Hence, prior studies typically separate adversarial robustness from the study of domain generalization and distribution shift.
We can unify adversarial robustness as a study of distribution/domain shift. Conceptually, we can define the shifted distribution in the adversarial setting as a distribution around the original distribution $\mathbb{P}_0$. We perform a perturbation on text $\boldsymbol{x} \in \mathbb{P}_0(\mathcal{X})$ to simulate OOD data from the shifted distribution. 
In other words, adversarial perturbations encode subtle domain shifts, as discussed in \S \ref{sec:diff}, and the shifted domains are characterized by the adversarial methods summarized in \S \ref{sec:adv_attacks}.

\subsection{Shifted Features}
To concretize the shifted features, we first define task-relevant features $s_{\text{rel}}$ and task-irrelevant features $s_{\text{ir}}$, 
which satisfy $\mathbb{P}(s_{\text{rel}}|y) \neq \mathbb{P}(s_{\text{rel}})$ and $\mathbb{P}(s_{\text{ir}}|y) = \mathbb{P}(s_{\text{ir}}$), respectively. $\mathbb{P}$ is an arbitrary distribution.

\paragraph{Background features and semantic features.}
When $\mathbb{P}$ is the ground-truth distribution $\mathbb{P}_\text{true}$, Arora et al., \citet{arora-etal-2021-types} name $s_{\text{rel}}$ as \textit{semantic features} since most realistic NLP tasks address the semantics of text, and $s_{\text{ir}}$ as \textit{background features}, which can be the syntax, writing style or task-irrelevant text.


\paragraph{Biased features are task-irrelevant features under $\mathbb{P}_\text{true}$ despite being learned as task-relevant features from $\mathbb{P}_0$.}
$\mathbb{P}_0(\mathcal{X})$ can be factorized into the conditional distributions of generalized features $g(\mathcal{X})$ and\textbf{ biased features $b(\mathcal{X})$}.
\begin{equation} 
\begin{split} \mathbb{P}_0(\mathcal{X}) = \sum_y \mathbb{P}_0(g(\mathcal{X})|y)\mathbb{P}_0(b(\mathcal{X})|y)\mathbb{P}_0(y),
\end{split}
\end{equation}
 assuming conditional independence between $b(\mathcal{X})$ and $g(\mathcal{X})$ given $y$.
By definition, $g(\mathcal{X})$ should always be task-relevant features in the true distribution $\mathbb{P}_\text{true}$, i.e., $g(\mathcal{X}) \in s_{\text{rel}}$, while biased features $b(\mathcal{X})$ are only predictive for ID dataset $\mathcal{D} \sim \mathbb{P}_0$. We can express $b(\mathcal{X})$ as $\{s|s \in \mathbb{P}_0 \text{ and } s \notin \Tilde{\mathbb{P}}\}$. 


\subsection{Shifted Features in Three OOD Types} \label{sec:diff}
This section shows how each OOD type can potentially shift the three kinds of features.
\paragraph{Shifted features on adversarial examples.}
Shifted features of adversarial examples depend on the attack methods. White-box and grey-box methods compute gradients or output logits, respectively, while black-box methods can only query final predictions.
With the guidance of gradients, 
white-box methods tend to shift the biased features since the input gradients can help identify the biased features for perturbations \citep{ilyas2019adversarial,wallace2019universal}. 
Perturbation types largely affect whether it is a semantic shift or background shift (See \S \ref{sec:adv_attacks}). For example, character editing may lead to unseen words, e.g., ``wonderful'' to ``wonderful'', which belongs to semantic shift.
In contrast, black-box methods explicitly change background features at the sentence level.
For example, Qi et al., \citet{qi-etal-2021-mind} paraphrase with different writing styles.

\paragraph{Shifted features on debiased data.}
Debiased data, by definition, is generated by removing biased features. 
However, the generation process often encodes either background features $s_{\text{ir}}$ or semantic features $s_{\text{rel}}$. For example, debiased data generated by syntactic templates \citep{mccoy-etal-2019-right} encode background shift, while MRC data requiring numerical reasoning to avoid dataset biases \citep{naik-etal-2018-stress} change semantic features.

\paragraph{Shifted features on NDS data.} 
We classify natural domain shift into two categories, namely genres and sub-populations.
Both background and semantic shifts can occur in both categories, such as different vocabularies associated with sub-populations or genres \citep{barrett-etal-2019-adversarial}. For example, for sentiment analysis, Tweets tend to express sentiments with emotions compared to the genre of movie reviews, and sub-populations can differ in the way to express emotions \citep{oprea2022sarcasm}.
Furthermore, DNNs can easily learn background features for each sub-population or genre as biased features.
Writing styles can also serve as another background feature that changes across genres \citep{arora-etal-2021-types}, e.g., text from fiction books to Wikipedia content.

\begin{table*}[!t]
    \centering
    \small
    \begin{tabular}{c|cc|cc}
        \toprule
        & \multicolumn{2}{c}{OOD Data Generation} &\multicolumn{2}{c}{ Evaluation} \\
          & Generated by & Required Knowledge &  Types & Metrics  \\
        \midrule
        Adversarial robustness  & Auto  & Model & Method-based & ASR; query \#  \\
        Adversarial transferability  & Auto/human  & Base model & Data-based & ID metrics  \\
        Domain generalization & Existing & Domain & Data-based & ID metrics  \\
        Dataset biases  & Auto/human & Biases/NA & Data-based & ID metrics \\
        \bottomrule
    \end{tabular}
    \caption{Comparison of OOD data generation and evaluation. ASR: attack success rate; query \#: number of queries; ID metrics: the same aggregate metrics for ID test data, e.g., accuracy for classifiers. }
    \label{tab:ood_generation}
\end{table*}

\section{OOD Performance Evaluation} \label{sec:evaluate_ood} 
This section compares the concrete methodologies of OOD data generation and evaluation protocols for each line of research. Their differences are summarized in Table \ref{tab:ood_generation}.

\begin{table*}[t!]
    \centering
    \small
\begin{tabular}{ llp{10cm} } 
 \toprule
 Biases & Tasks & Related Works \\
 \midrule
 \multirow{4}{*}{Lexical correlations} 
    & TC &  Geva et al., \citet{geva2019we} \\
    \cmidrule{2-3}
    & NLI & Hypothesis-only bias \citep{gururangan2018annotation} \\
    \cmidrule{2-3}
    & MRC & Question word matching \citep{sugawara-etal-2018-makes} \\
    \cmidrule{2-3}
    & Fact checking & Claim-only bias \citep{schuster-etal-2019-towards} \\
    \cmidrule{2-3}
    & Machine Translation & Word disambiguation bias \citep{emelin-etal-2020-detecting} \\
    \hline
 
  \multirow{2}{*}{Lexical overlap} 
    & NLI  & Premise-hypothesis overlap \citep{mccoy-etal-2019-right} \\
    \cmidrule{2-3}
    & MRC & Context matching \citep{sugawara-etal-2018-makes} \\
    \cmidrule{2-3}
    & Paraphrase  & Zhang et al., \citet{zhang-etal-2019-paws} \\
    \hline
    Positional bias & Summarization & Kedzie et al., \citet{kedzie-etal-2018-content} \\
    \hline
 \bottomrule
\end{tabular}
\caption{Dataset biases for various NLP tasks. TC: Text Classification; NLI: Natural Language Inference; MRC: Machine Reading Comprehension; Paraphrase: Paraphrase Identification.}
\label{tab:dataset_biases}
\end{table*}

\begin{table*}[t!]
    \centering
    \small
\begin{tabular}{ llp{10cm} } 
 \toprule
 Methods & Tasks & Debiased Datasets \\
 \midrule
 \multirow{4}{*}{Biased inspired} 
    & TC &  c-IMDB \citep{Kaushik2020Learning} \\
    \cmidrule{2-3}
    & NLI & HANS \citep{mccoy-etal-2019-right}, Stress Text \citep{naik-etal-2018-stress}  \\
    \cmidrule{2-3}
    & MRC & Para-SQUAD \citep{lai-etal-2021-machine} Adv-SQUAD \citep{jia-liang-2017-adversarial} \\
    \cmidrule{2-3}
    & Paraphrase & PAWS \citep{zhang-etal-2019-paws} \\
     \cmidrule{2-3}
    & Fact Checking & FEVER-Symmetric \citep{schuster-etal-2019-towards} \\
    \hline
 
  \multirow{3}{*}{Systematic} 
    & NLI  &  SNLI-AFLITE \citep{pmlr-v119-bras20a} \\
    \cmidrule{2-3}
    & Reasoning & Winogrande \citep{sakaguchi2020winogrande} \\
    \cmidrule{2-3}
    & CGI  & Swag \citep{zellers2018swag} \\
 \bottomrule
\end{tabular}
\caption{Debiased datasets for various NLP tasks. TC: Text Classification; NLI: Natural Language Inference; MRC: Machine Reading Comprehension; Paraphrase: Paraphrase Identification; Reasoning: Commonsense Reasoning; CGI: Commonsense Grounded Inference.}
\label{tab:debiased_datasets}
\end{table*}

\subsection{NDS Data Generation}
NDS data can be generated from different genres or sub-populations. 
The former can be acquired from various data sources, whereas the latter involves partitioning data into sub-populations based on attributes of entities or individuals.
Diverse sets of data genres may display a fusion of unique textual styles, syntactic structures, and lexicons. 

\paragraph{Genres.}
Different genres refer to text written by different authors or annotators or for different audiences, or come from different data sources.
For example, academic papers are written in formal language while online reviews on Yelp contain non-standard orthography. 
Since a training dataset typically comes from a specific data source and contains only one genre, even well-trained models generalize poorly to different genres. Although some datasets are created to include multiple genres of text, models still perform worse beyond the coverage of these genres. 
For example, MultiNLI for NLI consists of text from ten distinct genres. However, even the large pretrained language models cannot generalize well to a different genre \citep{arora-etal-2021-types}, such as SNLI from image captions or WNLI from fiction books.
Take text classification as another example. Classifiers cannot perform well on text from disjoint annotators, e.g., SST data from experts v.s. lengthy IMDB reviews from laymen \citep{hendrycks-etal-2020-pretrained,arora-etal-2021-types}.
\citet{pmlr-v119-miller20a} find that MRC models trained on Wikipedia hardly generalize on data from New York Times articles, Reddit posts, and Amazon product reviews.
    
\paragraph{Sub-populations.}
Multiple studies identify NDS data by dividing data into sub-populations according to different attributes of objects or individuals obtained from metadata.
Arora et al., \citet{arora-etal-2021-types} select reviews from different businesses, e.g., restaurant reviews v.s. movie reviews. 
For the same business, Hendrycks et al., \citet{hendrycks-etal-2020-pretrained} split them according to different types of products (e.g., women's clothing, shoes) or restaurants (e.g., American v.s. Chinese restaurants).
Furthermore, model performance can vary on data with different demographics, which is closely connected to another line of research about fairness. For example, Borkan et al., \citet{borkan2019bias} find the worst test accuracy on non-toxic comments from the black population.

\subsection{Debiased Data Generation} \label{sec:dataset_bias}
In this section, we first divide dataset biases into two categories and introduce a text summarization dataset bias that does not fall under either one.
We then classify data-generating methods into two categories according to the knowledge of dataset biases.
Table \ref{tab:dataset_biases} and \ref{tab:debiased_datasets} summarize dataset biases and debiased datasets for various NLP tasks, respectively.  

\subsubsection{Dataset Biases}
\paragraph{Lexical correlations.} 
Words frequently appearing in the training examples of a particular class can be predictive biases for that class.
For example, when NLI annotators create a majority of contradictory hypotheses by negating premises, models learn the spurious correlation between negation words (`no', `never') and labels (\textit{hypothesis-only reliance} \citep{gururangan2018annotation}). The success of class-wise adversarial attacks comes from exploring these predictive words to some extent \citep{wallace2019universal}.
Similarly, claim-only models can perform well on ID test data without the context of evidence for fact-checking due to predictive words \citep{schuster-etal-2019-towards}.
MRC models can find the answer in a given paragraph by matching the question type, e.g., ``November 2014'' for a ``when'' question (question word matching \citep{lai-etal-2021-machine} or entity type matching \citep{sugawara-etal-2018-makes}), and hence they are easily distracted to the wrong answers on OOD data \citep{jia-liang-2017-adversarial}. Emelin et al., \citet{emelin-etal-2020-detecting} observe that machine translation systems have a tendency to disambiguate words based on the sense that occurs most frequently in the training data.

\paragraph{Lexical overlap.}
Lexical overlap is another predictive indicator for sentence-pair classification. 
For example, the high overlap between the premise and hypothesis leads to the ``entailment'' prediction from NLI models (premise-hypothesis overlap \citep{mccoy-etal-2019-right}). 
The classifiers for paraphrase identification also tend to predict highly overlapping sentences as paraphrases \citep{zhang-etal-2019-paws}. 
Besides, MRC models can locate the sentence with maximum overlap words in the paragraph (context matching \citep{sugawara-etal-2018-makes}) and then use question word matching to extract the correct answer.

\paragraph{Positional biases for text summarization.}
News or journal articles tend to summarize content in the lead paragraphs (positional bias or layout bias). Specifically, Kedzie et al., \citet{kedzie-etal-2018-content} show that 88.6\% reference summaries from the training examples of NYT come from the first half of documents (69\% for DUC, 71.7\% for CNN/DM). 

\subsubsection{Methodologies}
There are two types of methods to generate debiased data: 1) breaking the spurious correlations; 2) filtering the examples containing superficial patterns automatically without recognizing any specific dataset biases.

\paragraph{Bias-informed approaches.}
Many studies generate debiased data by reversing the correlations between dataset biases and labels for text classification.
For natural language inference (NLI), Mccoy et al., \citet{mccoy-etal-2019-right} specify linguistic phenomena (constituent and subsequent heuristics) behind lexical overlap. They devise heuristics-based syntactic templates to create sentence pairs that have high lexical overlap but contradict each other.
Kaushik et al., \citet{Kaushik2020Learning} construct counterfactual examples by breaking spurious correlations in sentiment analysis datasets (i.e., IMDB) and NLI datasets.
They annotate the text into the targeted label by largely keeping the original text. In this way, the dataset biases of the original label are kept.
To break the lexical overlap between paraphrase pairs, Zhang et al., \citet{zhang-etal-2019-paws} create non-paraphrase sentences via word scrambling. This method changes the meanings of paraphrases but keeps the overlapping words.
For MRC, Jia and Liang \citet{jia-liang-2017-adversarial} add a sentence containing words overlapping with the question into paragraphs and find that models tend to select answers from the sentence. They also combine the bias-inspired approach with a grey-box attack to generate adversarial examples.
Gardner et al., \citet{gardner-etal-2020-evaluating} rely entirely on experts who have the knowledge of dataset biases to generate debiased data. 

\paragraph{Systematic approaches.}
Debiased data can also be generated by bias mitigation techniques without knowing dataset biases.
For example, Bras et al., \citet{pmlr-v119-bras20a} use a simple classifier to identify biased examples and generate debiased datasets for NLI, commonsense reasoning and grounded commonsense inference, respectively.
Although some works try to train robust models with such filtered data, filtering always leads to significant drops on ID data and even OOD data \citep{gokhale-etal-2022-generalized}. Therefore, it can be more practical to use them for OOD evaluation.


\subsection{Adversarial Example Generation} \label{sec:adv_attacks}
Given a well-trained model, adversarial examples are generated by modifying the given text (or reference data) to make the model output wrong predictions. 
Typically, the generation process satisfies semantics-preserving and label-preserving assumptions.
Mostly, semantics preservation can guarantee the same label for a perturbed text. Thus, a robust model should generate invariant outputs $\underset{y*}{\arg \max} f(\boldsymbol{x})=\underset{y*}{\arg \max} f(\Tilde{\boldsymbol{x}})=y$ under semantics-preserving perturbations, where $y$ is the ground-truth label and $\Tilde{\boldsymbol{x}}$ is the perturbed text.
Adversarial attacks have been widely studied to automatically generate adversarial examples. We categorize the attack methods according to their perturbation types, perturbation space and adversary's knowledge. 
Table \ref{tab:categorizations_attack} exemplifies typical methods for each category.

\begin{table*}[t!]
    \centering
    \small
\begin{tabular}{ llp{10cm} } 
 \toprule
 Knowledge & Perturbation Space & Perturbation Types \\
 \midrule
 \multirow{2}{*}{White-box} 
    & Continuous &   Word substitution \citep{gong2018adversarial}, UAP \citep{song2020universal}, Character editing \citep{liu-etal-2022-character}\\
    \cmidrule{2-3}
    & Discrete & Word substitution \citep{liang2017deep,ebrahimi2017hotflip}, UAP \citep{wallace2019universal,behjati2019universal}\\
    \midrule
    
   Grey-box  
    & Discrete & Word substitution \citep{alzantot2018generating,zang2019word,jin2020bert,li-etal-2020-bert-attack}, 
    Character editing \citep{gao2018-deepwordbug} \\
    \midrule
 
  \multirow{2}{*}{Black-box} 
    & Continuous &   Paraphrasing \citep{zhao2017generating} \\
    \cmidrule{2-3}
    & (Only generation model) & Paraphrasing \citep{iyyer-etal-2018-adversarial,ribeiro-etal-2018-semantically,qi-etal-2021-mind}\\
    \midrule
 \bottomrule
\end{tabular}
\caption{Classification of Attack Methods Based on Adversaries' Knowledge, Perturbation Space, and Perturbation Types. UAP refers to Universal Adversarial Perturbation (UAP). Several studies employ generation models to produce paraphrases without adding any noise for perturbations. }
\label{tab:categorizations_attack}
\end{table*}

\subsubsection{Perturbation Types}
There are four common perturbation types that exhibit different granularities, namely, character editing, word substitution, paraphrasing and universal adversarial perturbation (UAP).

Character editing mimics real-life accidental typos or spelling variants in social media via character swapping (e.g., ``place'' $\Rightarrow$ ``plcae'', deletion (e.g., ``artist'' $\Rightarrow$ ``arlist''), insertion (e.g., ``computer'' $\Rightarrow$ ``comnputer'') and substitution (e.g., ``computer'' $\Rightarrow$ ``computor'') \citep{gao2018-deepwordbug}. 
Humans are robust to spelling errors maintaining certain morphological or phonological characteristics, e.g., ``computer''. Hence, a few edits on a word would not affect the human perception of the word but may lead to completely opposite predictions by the models.
In contrast, word substitutions require semantics-preserving constraints on substitute words or rely on a vocabulary of synonyms for substitutions \citep{zang2019word}.
Paraphrasing is seldom considered in adversarial attacks due to the difficulty of generating high-precision paraphrases. 

The last type of adversaries follows the work to generate UAPs for images, which perturb any images of a particular class \citep{moosavi2017universal}.
Behjati et al., \citet{behjati2019universal} generate the UAPs for NLP models in the form of n-grams (i.e., non-sensical phrases), which lead to misclassification when appended to text of a class. Wallace et al., \citet{wallace2019universal,wallace2020imitation} even extend the idea of UAPs for MRC, language modeling and machine translation.
These studies assume that inserting non-sensical phrases does not change 
the semantics of text and ground-truth labels. Song et al., \citet{song2020universal} generate natural phrases (e.g., ``natural energy efficiency'') as UAPs via Adversarially Regularized Auto Encoder (ARAE). However, it can only maintain the coherence of UAPs rather than the whole perturbed text. 

\subsubsection{Perturbation Space}
A text consists of a sequence of tokens (e.g., words or subwords), denoted as $\boldsymbol{x}={w_1, w_2, \ldots, w_T }$, where $T$ is the number of tokens in the vocabulary $\mathcal{V}=\{w_1, w_2, \ldots, w_K\}$ ($K$ is the size of the vocabulary). 
Note that the vocabulary can differ from the ones of the victim models, especially under black-box attacks.

All types of perturbations, except for paraphrasing, can be perturbed in a discrete space. When perturbing via word substitutions or UAPs, the objective is to identify significant tokens in $\boldsymbol{x}$ and substitute them with tokens from the vocabulary $\mathcal{V}$ in order to generate incorrect model predictions. This is a combinatorial optimization problem, where the size of the search space is $K^T$. The search space is intractable and is typically addressed using heuristic-based approaches \citep{alzantot2018generating,zang2019word} or approximation methods \citep{ebrahimi2017hotflip,wallace2019universal}.
Character editing also requires identifying significant tokens, but it involves a subsequent manipulation of characters within the identified tokens \citep{gao2018-deepwordbug}. Assuming the average token length is $l$, the search space becomes $l^T$, leading to combinatorial explosion of modifications.

There are two types of methods to perturb text in the continuous space: 
1) perturbing tokens (characters/subwords/words) in the embedding space. Gong et al., \citet{gong2018adversarial} add continuous perturbations directly to the continuous representation $e_{w}$ of token $w$ via element-wise addition between the perturbation $\eta$ and $e_{w}$, similar to pixel-wise addition for images \citep{goodfellow2014explaining}.
To transform $\eta + e_{w}$ back to the text, they search for the nearest token $\Tilde{w}$ in the embedding space where $ e_{\Tilde{w}}+\eta \in \mathcal{V}$. However, it probably generates a semantically variant token, because the closest and most legible token can be far away from the original one in projection space. Note that they still require searching for important tokens.
2) reparameterization for paraphrasing or UAPs. Zhao et al., \citet{zhao2017generating} reparameterize the text $\boldsymbol{x}$ into $\boldsymbol{z}$ in a continuous space and then perturb $\boldsymbol{z}$ into $\Tilde{\boldsymbol{z}}$. They also train a text generator (e.g., LSTM) to decode $\Tilde{\boldsymbol{z}}$ back to text.
Song et al., \citet{song-etal-2020-adversarial} apply the reparameterization trick for UAP generation.

Moreover, generation models are commonly used to create paraphrases such as machine translation models for back-translation \citep{ribeiro-etal-2018-semantically},
a syntactically controlled paraphrase network \citep{iyyer-etal-2018-adversarial} or a text style transfer model \citep{qi-etal-2021-mind}. The latter two specifically change task-irrelevant features.


\subsubsection{Adversary's Knowledge}
We can categorize the attack methods into three types (white-box, grey-box and black-box) according to the three levels of model knowledge: model parameters, output logits (i.e., estimated probability distribution) and final predictions. 

The white-box methods require model parameters to propagate gradients back to the input (e.g., the gradient of an adversarial loss with respect to the input word embedding $\nabla_{e_w} \mathcal{L}$). They need $\nabla_{e_w} \mathcal{L}$ to find important tokens and generate perturbations, e.g., the noise in the embedding space or substitute tokes. 
Liang et al., \citet{liang2017deep} adapt the gradient magnitude to find important words for perturbations and search for substitute words for 
each targeted label since the magnitude in each dimension of the word embedding indicates the sensitiveness of the prediction to the change.
Gong et al., \citet{gong2018adversarial} directly use gradients to perform perturbations on the embedding space.
Ebrahimi et al., \citet{ebrahimi2017hotflip} and Wallace et al., \citet{wallace2019universal,wallace2020imitation} use the gradients and word embeddings to approximate the loss for word substitutions. Specifically, they approximate the loss change of substituting a word $w$ with another word $s$ in the vocabulary by the inner product of 
the word embedding $e_s$ and $\nabla_{e_w} \mathcal{L}$, where $\mathcal{L}$ is the adversarial loss. The selected word $s$ is expected to minimize the adversarial loss on the perturbed text $\Tilde{\boldsymbol{x}}$.
\begin{equation} \label{eq:unlearnable_approx}
\begin{split} 
    \underset{s}{\arg \min } \quad & \mathcal{L}(\Tilde{\boldsymbol{x}}, y) \\
      \approx \quad & e_s^{\mathrm{T}} \nabla_{e_w} \mathcal{L}(\boldsymbol{x}, y)
\end{split}
\end{equation}

Under the grey-box setting, an adversary only has access to the probability distribution (or logits) over the output variable $Pr(y|\boldsymbol{x})=f(\boldsymbol{x})$, where $Pr(y_i|\boldsymbol{x})$ is the probability of the outcome $y_i$ given the text $\boldsymbol{x}$.
Grey-box methods employ the probabilistic information to measure how important each token $w_i \in \boldsymbol{x}$ is to the prediction.
For $w_i$, its importance score $\boldsymbol{I}_i$ is commonly calculated as the difference of the probabilities of the correct class
$Pr(y_\text{true}|\boldsymbol{x}) - Pr(y_\text{true}|\boldsymbol{x}_{\backslash w_i})$ \citep{li-etal-2020-bert-attack}, where
$x_{\backslash w_i}$ can either be $\{w_1, \ldots, w_{i-1}, w_{i+1}, \ldots, w_T\}$ (delete $w_i$) \citep{li2019textbugger,jin2020bert} 
or $\{w_1, \ldots, \textit{UNK}, \ldots, w_T\}$ (replace $w_i$ with the unknown token \textit{UNK}) \citep{gao2018-deepwordbug,li-etal-2020-bert-attack}.
Jin et al., \citet{jin2020bert} also add the change to the misclassified class $ f_{\Tilde{y}}(\Tilde{x}) - f_{\Tilde{y}}(x)$ if $\underset{y*}{\arg \max} f(\Tilde{x}) \neq \underset{y*}{\arg \max} f(x)$, where $\Tilde{y}$ is the misclassified class, and $f_{\Tilde{y}}(\Tilde{x})$ is the probability of the misclassified class for the perturbed sample. 
Afterward, grey-box methods apply heuristics to search for substitute words or evaluate noisy words from character editing via $Pr(y_\text{true}|\Tilde{\boldsymbol{x}})$. 

The black-box attacks require no information about the victim model and commonly generate paraphrases. We have discussed the concrete methods in the previous paragraphs for perturbation types and space.

\subsection{Evaluation Protocols}
\paragraph{Data-based evaluation.}
There are different evaluation protocols for the three data types.
NDS data and debiased data are evaluated on the same aggregate metrics for ID data (e.g., accuracy for text classification and F1 scores for MRC). 
The drops of these metrics from ID data to OOD data reveal the model's ability of OOD generalization.

\paragraph{Method-based evaluation.}
It is meaningless to apply the ID metrics to adversarial examples unless used for adversarial transferability, since adversarial robustness only concerns the misclassified examples for the victim model. 
Instead, adversarial robustness is evaluated by the effectiveness of an attack method generating adversarial examples. The common evaluation metrics are \textit{attack success rate} and \textit{the number of queries}. 
Specifically,
given a test sample $\mathcal{D}_{\text{test}}$, \textit{attack success rate} measures the proportion of the data points in $\mathcal{D}_{\text{test}}$ which can be perturbed to fool a victim model. It is calculated by dividing the number of adversarial examples that successfully achieve attack goals, e.g., misclassification, by the size of $\mathcal{D}_{\text{test}}$.
\textit{The number of model queries} during an attack measures how many perturbed examples have been evaluated before a valid adversarial example appears. Besides the query for predictions, white-box and grey-box methods also query victim models to get gradients or output logits.
These two metrics can be used to compare the adversarial robustness between models (with fixed attack methods) \citep{gokhale-etal-2022-generalized} and the performance of different attack methods \citep{li-etal-2020-bert-attack} (with fixed victim models). 

\paragraph{Adversarial transferability.} 
Adversarial transferability is another evaluation type for adversarial robustness but uses the data-based evaluation method. It generates adversarial examples from a base model and then uses them to evaluate other models. 
The vulnerability can transfer across different model architectures and datasets \citep{wallace2020imitation}. Liang et al., \citet{liang2020does} show that transferability can even occur across tasks as long as the source data distribution and the target one are close. Adversarial transferability encourages the generation of costly human-made adversarial examples since they can be helpful in evaluating and improving the robustness of various models.
For example, Nie et al., \citet{nie-etal-2020-adversarial} and Wallace et al., \citet{wallace2019-trick} show model predictions or interpretations (e.g., highlighting words that are important for model predictions) to crowdworkers to assist in the generation of adversarial examples. 

\section{Opportunities and Challenges} \label{sec:challenges}
This section provides insights for future work based on the above OOD discussions. The opportunities and challenges related to adversarial robustness are also investigated.

\subsection{General Challenges and Opportunities}
Our study provides insights into OOD evaluation, OOD generalization and OOD detection.

\paragraph{A benchmark for all the OOD types.}
Developing a comprehensive benchmark for all the OOD types can help evaluate the model's actual ability of language understanding and identify potential incapacity.
Although there are open-source tools developed for OOD evaluation, most of them can only evaluate one perspective of the distribution shift.
It is challenging to develop such a benchmark due to the idiomatic lock-in in each line of research, different data-generating processes and evaluation protocols.
The disclosure of their connections and differences in this survey can provide insights into developing such a comprehensive benchmark for OOD evaluation.
For example, adversarial robustness can be evaluated via adversarial transferability to unify evaluation metrics. 

\paragraph{Data-generating processes may break the assumptions of covariate shift.}
Although all the three research lines aim to evaluate data with covariate shifts, some processes may break the assumptions of covariate shifts, i.e., fixed $\mathbb{P}_0(\mathcal{Y}|\mathcal{X})$ and $\mathbb{P}_0(\mathcal{Y})$. These two assumptions also indicate $\mathbb{P}_0(\mathcal{X}|\mathcal{Y})$ as a special case of covariate shift due to $\mathbb{P}_0(\mathcal{X}|\mathcal{Y})=\frac{\mathbb{P}_0(\mathcal{Y}|\mathcal{X})\mathbb{P}_0(\mathcal{X})}{\mathbb{P}_0(\mathcal{Y})}$.
For example, although $\mathbb{P}_0(\mathcal{Y}|\mathcal{X})$ is always assumed the same for most NLP tasks since humans usually annotate each sentence with the same label, gradient-based methods may generate adversarial examples with labels mismatching with humans' perceptions.
In addition, 
some methods for generating debiased data may change $\mathbb{P}_0(\mathcal{Y})$. 
For example, Zhang et al., \citet{zhang-etal-2019-paws} balance examples with high lexical overlap for all kinds of labels in their debiased dataset to prevent the feature of lexical overlap from being a predictive indicator for paraphrase identification.
Future work that evaluates covariate shifts (i.e., the changes of input domains) may better consider the two assumptions for data-generating processes and analyze their effects.

\paragraph{Enhance general OOD performance.}
The three types of out-of-distribution (OOD) data can be created by altering task-irrelevant features, as explained in Section \ref{sec:define}. Our first argument suggests that OOD generalization techniques should aim to improve the acquisition of general linguistic or semantic knowledge to perform well on all three types of OOD evaluation. This accounts for the observation that unsupervised pretraining and data augmentation strategies for learning general linguistic or semantic knowledge can enhance generalization on different domains and adversarial robustness \citep{gururangan-etal-2020-dont,gokhale-etal-2022-generalized}. Notably, such augmentation techniques \citep{wei-zou-2019-eda} may solely modify the background features of training samples without any adversarial process.
Our second argument is that even though dataset filtering for bias mitigation may improve model performance on biased datasets, it may lead to inferior performance on NDS data and decreased adversarial robustness , as biased data still contains knowledge of semantic and background features that can enhance general OOD performance. Supporting evidence for this argument is provided by Bras et al., \citet{pmlr-v119-bras20a} and Gokhale et al., \citet{gokhale-etal-2022-generalized} across various NLP tasks.

\paragraph{Utilizing detection methods across different research lines.}
The connections between OOD types regarding the shifted features also motivate future researchers to apply OOD detection methods across the three research lines. For example, since density estimation methods can effectively detect NDS data with background shift \citep{arora-etal-2021-types}, they may defend against black-box attacks, which always generate adversarial examples by shifting background features.

\subsection{Challenges and Opportunities for Adversarial Robustness}
\paragraph{Beyond semantics preservation.}
Adversarial examples can be generated beyond the assumption of semantics preservation.
Chen et al., \citet{chen2022balanced} define obstinate adversarial examples, which satisfy two conditions: 1) their ground-truth labels after the perturbation are changed, but 2) the victim model maintains the original predictions. Similarly, Kaushik et al., \citet{Kaushik2020Learning} minimally perturb test examples to change the labels of test examples.
There are many reasons to explore obstinate adversarial examples.
Gardner et al., \citet{gardner-etal-2020-evaluating} argue that such examples characterize the correct decision boundary for the task. 
Chen et al., \citet{chen2022balanced} find that models with adversarial training on these examples reveal vulnerability to obstinate adversarial examples. 
Also, breaking the assumption of semantics preservation allows more adversarial behaviours for tasks beyond single-sentence classification. For example, Song et al., \citet{song-etal-2020-adversarial} generate nonsensical or natural sentences, leading to invariant model outputs for sentence-pair classification. 
Wallace et al., \citet{wallace2020imitation} change malicious content into nonsensical text which still makes machine translation models translate it into bad language. 

\paragraph{Developing realistically harmful adversarial behaviours.}
There are different model behaviours designed for malicious intents during adversarial attacks.
For classification tasks,
we can either specify the expected prediction (targeted attack) or accept any prediction different from the correct label (non-targeted attack). The harm of attack depends on the specific task or labels. For example, it is harmful to generate adversarial examples to evade the detection of misinformation or toxic content.
In contrast, there are many possibilities of adversarial behaviours for generation tasks due to various combinations of input and output text. 
Cheng et al., \citet{cheng2020seq2sick} define an attack where all the words in the output text are different from the original output sequence while the input is similar.
This kind of attack can lower the standard evaluation metrics based on n-grams, like BLEU scores. 
Wallace et al., \citet{wallace2020imitation} develop UAPs to make MT models hardly generate any translation or output random words. Targeted keyword attacks \citep{cheng2020seq2sick} can make models generate targeted words. Future researchers can define more realistically harmful behaviours, such as the generation of malicious nonsense or racial materials.

\paragraph{Adversarial robustness may not be a good proxy for realistic scenarios.} 
Adversarial perturbations may make text deviate from their real-world distribution and generate rarely occurred examples. Particularly, pure gradient-based methods only pursue worst-case perturbations without considering the naturalness of adversarial examples. 
Hence, to alleviate this problem, adversarial candidates are commonly validated by some metrics such as language model perplexity \citep{alzantot2018generating}, and
part-of-speech matching \citep{ebrahimi2017hotflip}. However, attack processes become computationally expensive by including these separate modules and rejecting most of the perturbed examples. 
Another type of approach is to use limited search space for adversarial perturbations, e.g., using synonyms for word substitutions with the compromise of worst-case performance \citep{zang2019word}.
It is worth exploring the attack methods that can inherently ensure the naturalness of perturbed text.
There are some scenarios, as summarized below, where adversarial robustness against rare and unnatural text is undoubtedly critical. 1) When models' behaviours can cause devastating outcomes, e.g., astronautics or legal services.
2) When models are deployed in ubiquitous, unforeseeable scenarios.
For example, 
Facebook translation once made the mistake of translating the simple phrase ``good morning'' in Arabic into ``attack it'' in English. 
Such a mysterious translation convinced the police that the user posting this message might launch a vehicle attack, resulting in the arrest of the user.
3) When attackers have sufficient malicious intent to explore rare cases, such as fact verification 
and security tasks where attackers would like to evade the model's detection to get unqualified access, e.g., propagating anti-social content or fake news on social media.


\section{Conclusion}
In this paper, we characterized and summarized the three mainstream lines of research on dataset biases, domain generalization and adversarial robustness. 
We encouraged future researchers to think comprehensively about OOD evaluation and improve the OOD generalization of NLP models on all the three types of OOD data.
We also highlighted the gap between adversarial robustness and realistic OOD evaluation.

\bibliographystyle{named}
\bibliography{ijcai23}

\end{document}